\documentclass[sigconf, authorversion]{acmart}

\AtBeginDocument{%
  \providecommand\BibTeX{{%
    \normalfont B\kern-0.5em{\scshape i\kern-0.25em b}\kern-0.8em\TeX}}}
\usepackage{natbib}
\usepackage{enumitem}
\usepackage{hyperref}
\hypersetup{
    colorlinks=true,
    linkcolor=blue,
    filecolor=magenta,      
    urlcolor=cyan,
    pdftitle={Overleaf Example},
    pdfpagemode=FullScreen,
    }

\usepackage{graphicx}

\copyrightyear{2021}
\acmYear{2021}
\setcopyright{acmcopyright}\acmConference[CIKM '21]{Proceedings of the 30th ACM International Conference on Information and Knowledge Management}{November 1--5, 2021}{Virtual Event, QLD, Australia}
\acmBooktitle{Proceedings of the 30th ACM International Conference on Information and Knowledge Management (CIKM '21), November 1--5, 2021, Virtual Event, QLD, Australia}
\acmPrice{15.00}
\acmDOI{10.1145/3459637.3482029}
\acmISBN{978-1-4503-8446-9/21/11}

\begin{document}

\title{Adversarial Robustness of Deep Learning: \\Theory, Algorithms, and Applications}


\author{Wenjie Ruan}
\affiliation{%
  \institution{University of Exeter}
  \city{Exeter}
  \country{UK}
  \postcode{78229}}
\email{w.ruan@exeter.ac.uk}

\author{Xinping Yi}
\affiliation{%
  \institution{University of Liverpool}
  \city{Liverpool}
  \country{UK}}
\email{xinping.yi@liverpool.ac.uk}

\author{Xiaowei Huang}
\affiliation{%
  \institution{University of Liverpool}
  \city{Liverpool}
  \country{UK}}
\email{xiaowei.huang@liverpool.ac.uk}


\begin{abstract}
 

This tutorial aims to introduce the fundamentals of adversarial robustness of deep learning, presenting a well-structured review of up-to-date techniques to assess the vulnerability of various types of deep learning models to adversarial examples. This tutorial will particularly highlight state-of-the-art techniques in adversarial attacks and robustness verification of deep neural networks (DNNs). We will also introduce some effective countermeasures to improve robustness of deep learning models, with a particular focus on adversarial training. We aim to provide a comprehensive overall picture about this emerging direction and enable the community to be aware of the urgency and importance of designing robust deep learning models in safety-critical data analytical applications, ultimately enabling the end-users to trust deep learning classifiers. We will also summarize potential research directions concerning the adversarial robustness of deep learning, and its potential benefits to enable accountable and trustworthy deep learning-based data analytical systems and applications.
\end{abstract}

\ccsdesc[500]{Computing methodologies~Artificial intelligence}




\maketitle

\section{Rationale} 

In recent years, we witness significant progress has been made in deep learning, which can achieve human- or superhuman-level performance on various data analytical tasks, such as image data recognition \cite{russakovsky2015imagenet}, natural language processing \cite{collobert2011natural}, and medical data analysis \cite{esteva2019guide}. Given the prospect of a broad deployment of DNNs in a wide range of applications, concerns regarding the safety and trustworthiness of deep learning have been recently raised \cite{goodfellow2018making,xu2020towards}. There is significant research that aims to address these concerns, with many publications appearing since the year of 2014 \cite{huang2020survey}. As we seek to deploy deep learning systems not only on virtual domains, but also in real systems, it becomes critical that a deep learning model can obtain satisfactory performance, but which are truly robust and reliable. Although many notions of robustness and reliability exist in different communities, one particular topic in machine learning community that has attract enormous attention in recent years is the adversarial robustness of deep learning: a deep learning model is fragile or extremely non-robust to an input that is adversarially perturbed, and such perturbations usually are invisible or insensible by humans \cite{goodfellow2018making}. This is although a very specific notion of robustness in general, but one that concerns the safety and trustworthiness of modern deep learning systems.

This tutorial seeks to provide a broad, hands-on introduction to the topic concerning adversarial robustness: {\em the widespread vulnerability of state-of-the-art deep learning models to adversarial misclassification} (i.e., adversarial examples). The goal is to combine both formal mathematical treatments and practical tools and applications that highlight some of the key methods and challenges concerning the adversarial robustness. This tutorial will specifically concentrate on three major research progress on this direction - adversarial attacks, defences and verification. As detailed in Section 5, some tutorials regarding this emerging direction already appeared in flagship conferences in machine learning, AI and computer vision communities, including IJCAI 2021\footnote{Towards Robust Deep Learning Models: Verification, Falsification, and Rectification in IJCAI 2021 (https://tutorial-ijcai.trustai.uk/)}, ECML/PKDD 2021\footnote{https://tutorial-ecml.trustai.uk/}, ICDM 2021\footnote{\url{https://tutorial.trustdeeplearning.com/}}, CVPR 2020, KDD 2019\footnote{Recent Progress in Zeroth Order Optimization and Its Applications to Adversarial Robustness in Data Mining and Machine Learning, in KDD 2019, CVPR 2020}, etc. Our tutorial is fundamentally different to those existing ones, which is i) {\bf more comprehensive}: we not only cover adversarial attacks but particularly concentrate on verification-based approaches which is able to establish formal robustness guarantees; ii) {\bf more application-oriented}: in the second part of our tutorial, we emphasize one particular defence technique that can significantly improve robustness of DNNs - adversarial training, which will shed a light on the development of robust deep learning models for real-world data analytical applications. The specific differences to each current similar tutorials are detailed in Section 5.

This tutorial can alarm the community to be aware of the safety vulnerabilities of deep learning on real-world data analytical solutions despite its appealing performance. We also envision, through this tutorial, AI and data mining researchers and engineers get a sense on how to evaluate the robustness of deep learning models (e.g., via adversarial attacks/perturbations and verification-based approaches) and how to design/train robust deep learning models (e.g., via defence, especially adversarial training). 

\section{Content details} 

\begin{itemize}[leftmargin=*]
    \item {\bf Introduction to adversarial robustness}: this part will introduce the concept of adversarial robustness by showing some examples from computer vision~\cite{zhang2021fooling}, natural language processing~\cite{jin2020bert}, medical systems~\cite{wu2021interpretable}, and autonomous systems~\cite{wu2021adversarial}. Specifically, we will demonstrate the vulnerabilities of various types of deep learning models to different adversarial examples. We will also highlight the dissimilarities of research focuses on adversarial robustness from different communities, i.e., attack, defense and verification.
    
    \item {\bf Adversarial attacks}: this part will detail some famous adversarial attack methods with an aim to provide some insights of why adversarial examples exit and how to generate adversarial perturbation effectively and efficiently. Specifically, we will present six typical adversarial attacks, including L-BFGS~\cite{szegedy2013intriguing}, FGSM~\cite{goodfellow2015explaining}, C\&W~\cite{carlini2017towards}, ZeroAttack~\cite{tu2019autozoom}, spatial-transformed attacks~\cite{xiao2018spatially}, universal attacks~\cite{zhang2020generalizing}. In the end of this part, we will also briefly introduce some adversarial attacks on other domains, including attacks on sentiment analysis systems~\cite{jin2020bert}, attacks on 3D point cloud models~\cite{hamdi2020advpc}, attacks on audio recognition systems~\cite{abdullah2019practical}.
    
    \item {\bf Adversarial defense}: this part will present an overview of state-of-the-art robust optimisation techniques for adversarial training \cite{madry2018towards}, with emphasis on distributional robustness and the interplay between robustness and generalisation. In particular, adversarial training with Fast Gradient Method (FGM)~\cite{goodfellow2015explaining}, Projected Gradient Method (PGM)~\cite{kurakin2016adversarial} will be introduced briefly, followed by the advanced methods promoting distributional robustness \cite{sinha2018certifying} from the viewpoints of robustness versus accuracy \cite{zhang2019theoretically}, supervised versus semi-supervised learning \cite{miyato2018virtual}, and the exploitation of local and global data information \cite{zhang2019defense,qian2021improving}. In addition, the interplay between robustness and generalisation will be discussed with respect to generalisable robustness and robust generalisation. A variety of regularisation techniques such as spectral normalisation~\cite{miyato2018spectral}, Lipschitz regularisation~\cite{virmaux2018lipschitz}, and weight correlation regularisation \cite{jin2020does} to promote generalisable robustness will be discussed, together with some recent advances to improve robust generalisation \cite{schmidt2018adversarially,yin2019rademacher,zhang2021towards}.
    
    \item {\bf Verification and validation}: this part will review the state-of-the-art on the verification techniques for checking whether a deep learning model is dependable. First, we will discuss verification techniques for checking whether a convolutional neural network is robust against an input, including constraint  solving based techniques (MILP, Reluplex~\cite{katz2017reluplex}), approximation techniques (MaxSens~\cite{xiang2018output}, AI$^2$~\cite{gehr2018ai2}, DeepSymbol \cite{10.1007/978-3-030-32304-2_15}), and global optimisation based techniques (DLV~\cite{huang2017safety}, DeepGO~\cite{ruan2018reachability,ruan2018global}, DeepGame~\cite{wicker2018feature,wu2020game}). Second, we will discuss some  software testing based methods which generate a large set of test cases according to coverage metrics, including e.g., DeepXplore~\cite{PCYJ2017}, and DeepConcolic~\cite{sun2018concolic,10.1145/3358233,berthier2021tutorials}, and their extension to recurrent neural networks \cite{9451178}. The dependability of a learning model can then be estimated through the test cases. Third, we will discuss how to extend these above techniques to work with a reliability notion which considers all possible inputs in an operational scenario \cite{10.1007/978-3-030-54549-9_16}. This requires the consideration of robustness and generalisation in a holistic way \cite{jin2020does,RAM2021}. Finally, we will summarize and outlook current state of this research field and future perspectives.
    
\end{itemize}

\section{Target audience and prerequisites}
This tutorial motivates and explains a topic of emerging importance for AI, and it is particularly devoted to anyone who is concerning the safety and robustness of deep learning models.  The target audience would be data mining and AI researchers and engineers who wish to learn how the techniques of adversarial attacks and verification as well as adversarial training can be effectively used in evaluating and improving the robustness of deep learning models. 
No knowledge of the tutorial topics is assumed. A basic knowledge of deep learning and statistical pattern classification is requested.

\section{Benefits}
Deep learning techniques now is not only pervasive in the community of computer vision and machine learning, but also widely applied on data analytical systems. For researchers and industrial practitioners who are developing safety-critical systems such as health data analytics, malware detection, and automatic disease diagnosis, the robustness and reliability of deep learning models are profoundly important. CIKM, as a flagship conference in data mining and knowledge management, has attracted huge amount of data scientists and engineers and many of them are using deep learning techniques. We envision, by presenting a tutorial concerning the robustness of deep learning at CIKM'21, the community can {\em i) be aware the vulnerability of deep learning models}, {\em ii) understand why such vulnerability exits in deep learning and how to evaluate its adversarial robustness}, and {\em iii) know how to train a robust deep learning model}. We believe, those mentioned goals are appealing to the audience in CIKM'21. In the meantime, a tutorial\footnote{Recent Progress in Zeroth Order Optimization  and Its Applications to Adversarial Robustness in Data Mining and Machine Learning, in CVPR2020, KDD 2019.} conerning similar topic already appeared in SIGKDD'19 and ICDM 2020, two flagship conferences in data mining. As such, we believe it is necessary and urgent to propose a more comprehensive tutorial concentrating this topic in CIKM'21 as well.

\section{Difference to similar tutorials}

\begin{itemize}[leftmargin=*]
    \item Rigorous Verification and Explanation of ML Models, in AAAI 2020. Link: \url{https://alexeyignatiev.github.io/aaai20-tutorial/}
\begin{itemize}
    \item The overlapping will be on the verification part, where the above tutorial only considers from the logic/binary perspective, while we will cover comprehensively constraint-solving based methods, approximation methods, and global optimisation methods. The other two topics of this tutorial, i.e., adversarial attack (Part-I) and defense (Part-III), are not covered in the above tutorial. 
\end{itemize}

\item Adversarial Machine Learning, in AAAI 2019, AAAI 2018. Link: \url{https://aaai19adversarial.github.io/index.html#org}
\begin{itemize}
\item The above tutorial focuses on adversarial attacks of classifier evasion and poisoning as well as the corresponding defense techniques, while this tutorial places the emphasis on the fundamentals of adversarial examples (Part I) and generalisable robust optimisation techniques for defense (Part III), which are not covered by the above tutorial. In addition, the topic of verification of this tutorial (Part II) is not covered at all in the above one.
\end{itemize}

\item Adversarial Machine Learning, in IJCAI 2018, ECCV 2018, ICCV 2017. Link: \url{https://www.pluribus-one.it/research/sec-ml/wild-patterns}
\begin{itemize}
    \item The above tutorial concentrates on demonstrating vulnerability of various machine learning models and the design of learning-based pattern classifiers in adversarial environments. Our tutorial, however, is primarily about adversarial robustness of deep neural networks, especially the safety verification (Part-II) and adversarial defense (Part-III) on DNNs are not covered by the above tutorial. The only overlapping will be the part of adversarial attacks, but ours is more comprehensive and deep learning-focused.
\end{itemize}

\item Adversarial Robustness: Theory and Practice, in NIPS 2018
\\Link: \url{https://adversarial-ml-tutorial.org/}
\begin{itemize}
    \item The above tutorial was given two years ago, concentrating on the verification-based approaches to establishing formal guarantees for adversarial robustness. Then it presents adversarial training and regularization-based methods for improving robustness. Our tutorial will be more up-to-date. The major differences are: 1) in adversarial attacks, we present more recent and advanced adversarial examples, such as universal and spatial-transformed one; 2) in verification, we are more comprehensive, except for the constraint-solving based methods, we also cover the approximation and global optimisation methods; and (3) in adversarial defense, our focus is on generalisable adversarial training via advanced spectral regularisation.
\end{itemize}

\item Recent Progress in Zeroth Order Optimization and Its Applications to Adversarial Robustness in Data Mining and Machine Learning, in CVPR 2020, KDD 2019. \\Link: \url{https://sites.google.com/view/adv-robustness-zoopt} 
\begin{itemize}
    \item The above tutorial concentrates on Zero-order optimization methods with a particular focus on black-box adversarial attacks to DNNs. Our tutorial is more comprehensive, which not only covers a wider range of adversarial attacks, but also presents verification approaches that can establish formal guarantees on adversarial robustness, as well as review state-of-the-art adversarial training methods that can defence adversarial attacks and improve DNN's robustness.
\end{itemize}

\end{itemize}

\section{Relevant Experience on the Topic}

\subsection{Relevant Tutorial Experience}

\begin{itemize}
    \item W. Ruan, X. Yi, X. Huang, Tutorial “Adversarial Robustness of Deep Learning Theory, Algorithms, and Applications”, The 20th IEEE International Conference on Data Mining (ICDM 2020), 17-20 Nov 2020, Sorrento, Italy\\
    Link: \url{https://tutorial.trustdeeplearning.com/}
    \item	W. Ruan, E. Botoeva, X. Yi, X. Huang, Tutorial "Towards Robust Deep Learning Models: Verification, Falsification, and Rectification", The 30th International Joint Conference on Artificial Intelligence (IJCAI 2021), 21-26 Aug 2021, Canada\\
    Link: \url{http://tutorial-ijcai.trustai.uk/}
    \item W. Ruan, X. Yi, X. Huang, Tutorial “Adversarial Robustness of Deep Learning Theory, Algorithms, and Applications”,  The 2021 European Conference on Machine Learning and Principles and Practice of Knowledge Discovery in Databases (ECML/PKDD 2021), 13-17 Sep 2021, Virtual\\
    Link: \url{http://tutorial-ecml.trustai.uk/}

\end{itemize}

\subsection{Citations of Relevant Works}

The following are the selected papers published by the presenters which are related to this tutorial\footnote{The citation numbers are from Google Scholar on 22 August 2021}.

\begin{itemize}
    \item[(1)] Safety verification of deep neural networks, CAV 2017, Citation = 623
    \item[(2)] Structural Test Coverage Criteria for Deep Neural Networks, in ACM Transactions on Embedded Computing Systems 2018, Citation = 196
    \item[(3)] Concolic Testing for Deep Neural Networks, in ASE 2018, Citation = 183
    \item[(4)] Feature-guided black-box safety testing of deep neural networks, in TACAS 2018, Citation = 162
    \item[(5)] Reachability Analysis of Deep Neural Networks with Provable Guarantees, in IJCAI 2018, Citation = 163
    \item[(6)] A survey of safety and trustworthiness of deep neural networks: Verification, testing, adversarial attack and defence, and interpretability, in Computer Science Review, 2020. Citation = 111
    \item[(7)] A game-based approximate verification of deep neural networks with provable guarantees, in Theoretical Computer Science, 2020. Citation = 61
    \item[(8)] Global Robustness Evaluation of Deep Neural Networks with Provable Guarantees for the Hamming Distance, in IJCAI 2019, Citation = 53
\end{itemize}

The overall citations of the above papers that are closely related to this tutorial are over {\bf 1,200} since 2017. The details are as below:

\begin{itemize}
    \item A seminal  paper (1), on the safety verification of deep learning, has attracted {\bf 600+} Google Scholar citations, which is one of the first papers on the verification of deep learning.
    
    \item A few papers including (5) (7) (8), on the verification of neural networks through global optimisation algorithms, have attracted {\bf 250+} citations. 
    
    \item A few papers including (2) (3) (4) (5), on the adversarial attacks of the robustness of neural networks, have attracted {\bf 500+} citations. 
    
    \item A recent survey paper (6), closely aligned with the topic of this tutorial, has attracted {\bf 100+} Google scholar citations since its publication in 2020. 
\end{itemize}

\section{Brief Resumes of Presenters}

\subsubsection{Dr Wenjie Ruan}
   Dr Wenjie Ruan is a Senior Lecturer of Data Science at University of Exeter, UK. Previously, he has worked at Lancaster University as a lecturer, and University of Oxford as a postdoctoral researcher. Dr Ruan got his PhD from University of Adelaide, Australia. His series of research works on {\em Device-free Human Localization and Activity Recognition for Supporting the Independent Living of the Elderly} have received {\em Doctoral Thesis Excellence} from The University of Adelaide. He was also the recipient of the prestigious DECRA fellowship from ARC (Australian Research Council). Dr Ruan has published 30+ top-tier papers in top venues such as AAAI, IJCAI, ICDM, UbiComp, CIKM, ASE, etc. His recent work on {\em reachability analysis on deep learning} is one of the most citable papers in IJCAI'18 (150+ citations since 2018), and his work on {\em testing-based falsification on deep learning} is also one of the most citable papers in ASE'18 (150+ citations since 2018). Dr. Ruan has served as Senior PC, or PC member for over 10 conferences including IJCAI, AAAI, ICML, NeurlPS, CVPR, ICCV, AAMAS, ECML-PKDD, etc. His homepage is: {\em \url{http://wenjieruan.com/}}.
    
 \subsubsection{Dr Xinping Yi} 
 Dr Xinping Yi is a Lecturer (Assistant Professor) of Electrical Engineering at the University of Liverpool, UK. He received his Ph.D. degree from Telecom ParisTech, Paris, France. Prior to Liverpool, he worked at Technische Universitat Berlin, Germany, EURECOM, France, UC Irvine, US, and Huawei Technologies, China. Dr Yi’s recent research lies in deep learning theory with emphasis on generalisation and adversarial robustness. Dr Yi has published 40+ papers in IEEE transactions such as IEEE Transactions on Information Theory (TIT), and machine learning conferences such as ICML, NeurIPS.
 He has served as programme committee members and reviewers at a number of international conferences and journals, such as ICML, ICLR, IJCAI, CVPR, ICCV, ISIT, ICC, TIT, Proceedings of IEEE, JSAC, TWC, Machine Learning.
 His homepage is: \url{https://sites.google.com/site/xinpingyi00/}

 \subsubsection{Dr Xiaowei Huang}
 
 Dr Xiaowei Huang is Reader of Computer Science at the University of Liverpool, UK. Prior to Liverpool, he worked at Oxford and UNSW Sydney. Dr Huang’s research concerns the safety and trustworthiness of autonomous intelligent systems. He is now leading a research group working on the verification, validation, and interpretability of deep neural networks. Dr Huang is the principle investigator of several Dstl projects concerning the safety and assurance of artificial intelligence, and the Liverpool lead on a H2020 project on the foundation of trustworthy autonomy and an EPSRC project on the security of neural symbolic architectures. He is actively engaged with both formal verification and artificial intelligence communities and has given a number of invited talks on topics related to the safety of artificial intelligence. Dr Huang has published 50+ papers in international conferences such as AAAI, IJCAI, CAV, TACAS, ASE, etc., and has served in program committees of 20+ international conferences. His homepage is: \url{https://cgi.csc.liv.ac.uk/~xiaowei/}

\paragraph{Acknowledgement}
Wenjie Ruan is supported by Offshore Robotics for Certification of Assets (ORCA) Partnership Resource Fund (PRF) on {\em Towards the Accountable and Explainable Learning-enabled Autonomous Robotic Systems (AELARS)} [EP/R026173/1].
\includegraphics[height=8pt]{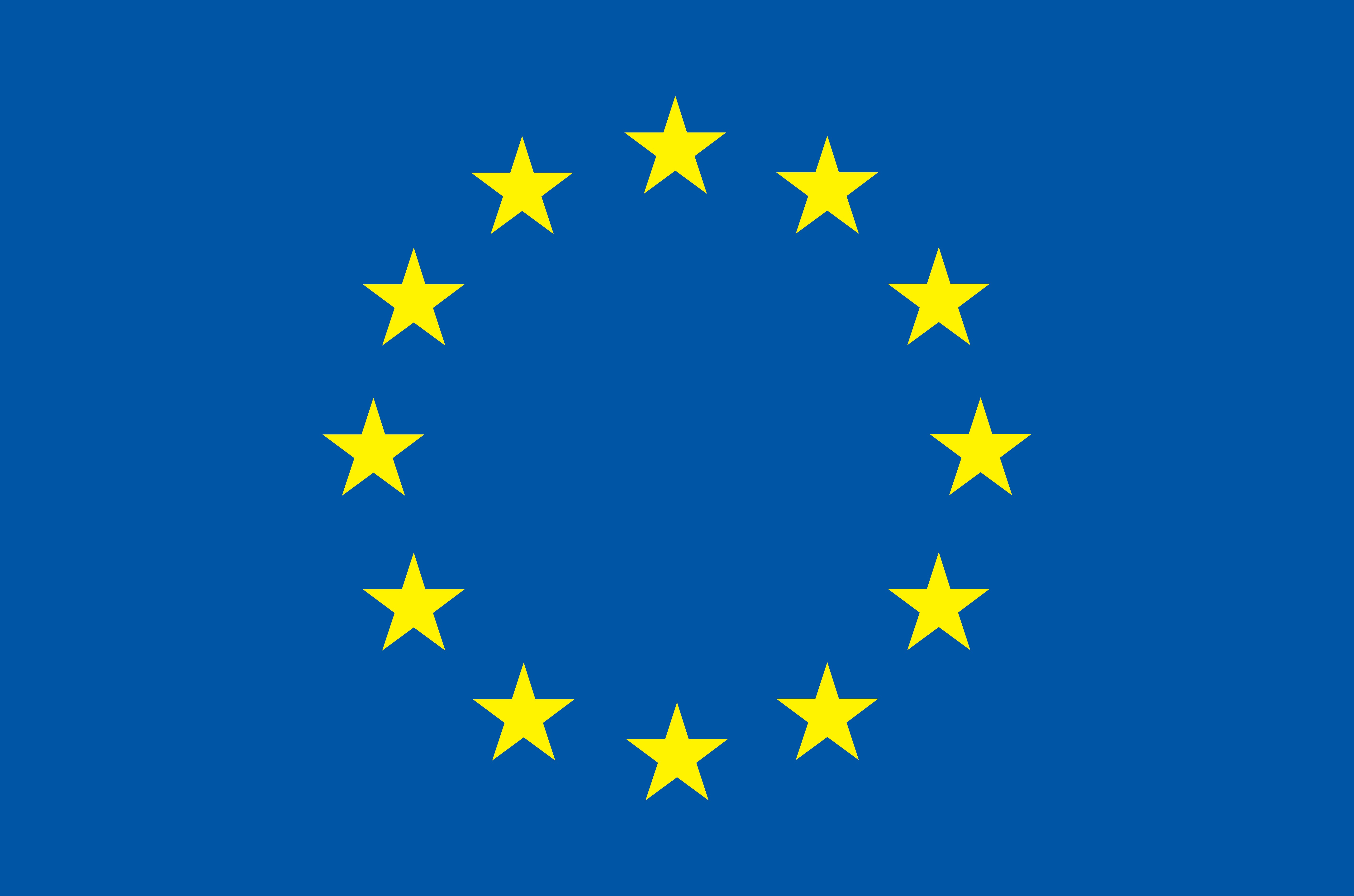}  XH has received funding from the European Union’s Horizon 2020 research and innovation programme under grant agreement No 956123, and is also  supported by the UK EPSRC (through the Offshore Robotics for Certification of Assets [EP/R026173/1] and End-to-End Conceptual Guarding of Neural Architectures [EP/T026995/1]).

 \small	
\bibliographystyle{ACM-Reference-Format}
\bibliography{sample-base}
\end{document}